\documentclass[12pt]{article}

\title{Multi-Weight Ranking for Multi-Criteria Decision Making\thanks{This paper is an extended version of an accepted contribution to the MODeM Workshop at ECAI  2023, Krakow, Poland, presented by D. Kostner.}}

\author{Andreas H. Hamel\footnote{Free University of Bolzano-Bozen, Faculty of Economics and Management, \href{mailto:andreas.hamel@unibz.it}{andreas.hamel@unibz.it}}, Daniel Kostner\footnote{Free University of Bolzano-Bozen, Faculty of Economics and Management, \href{mailto:daniel.kostner@unibz.it}{daniel.kostner@unibz.it}, corresponding author}
}
\date{{\small \today}}

\usepackage{array} 
\usepackage{verbatim} 
\usepackage{amsmath, amssymb, enumerate, color, graphicx}
\usepackage{stmaryrd}

\newtheorem{theorem}{Theorem}

\newtheorem{definition}[theorem]{Definition}
\newtheorem{proposition}[theorem]{Proposition}

\numberwithin{equation}{section}  
\numberwithin{figure}{section}    
\numberwithin{table}{section}     
\numberwithin{theorem}{section}

\newcommand{\cb}[1]{\ensuremath{ \left\{ #1 \right\} }}

\newcommand{\bs}{\backslash}

\newcommand{\R}{\mathrm{I\negthinspace R}}

\newcommand{\N}{\mathrm{I\negthinspace N}}

\newcommand{\triup}{{\rm \vartriangle}}
\newcommand{\trido}{{\rm \triangledown}}

\newcommand{\Int}{{\rm int\,}}

\usepackage[
colorlinks=true, linkcolor=blue, citecolor=blue, urlcolor=blue, filecolor=blue
auskommentieren und unten pdfborder aktivieren]{hyperref}

\definecolor{color0}{gray}{.5}
\definecolor{color1}{rgb}{0,.2,.8}
\definecolor{color2}{rgb}{1,.2,0}
\definecolor{color3}{rgb}{.2,.7,.6}

\begin{document}

\maketitle

\begin{abstract}
Cone distribution functions from statistics are turned into Multi-Criteria Decision Making tools. It is demonstrated that this procedure can be considered as an upgrade of the weighted sum scalarization insofar as it absorbs a whole collection of weighted sum scalarizations at once instead of fixing a particular one in advance. As examples show, this type of scalarization--in contrast to a pure weighted sum scalarization--is also able to detect ``non-convex" parts of the Pareto frontier. Situations are characterized in which different types of rank reversal occur, and it is explained why this might even be useful for analyzing the ranking procedure. The ranking functions are then extended to sets providing unary indicators for set preferences which establishes, for the first time, the link between set optimization methods and set-based multi-objective optimization. A potential application in machine learning is outlined.
\end{abstract}

{\bf Keywords.} multi-objective optimization, ranking function, cone distribution function, rank reversal, set optimization, supervised learning

\medskip
{\bf Mathematical Subject Classification.} 68T05, 68Q32, 90C29

\section{The ranking problem}

The Multi-Criteria Decision Making (MCDM) dilemma is that best alternatives are looked for among a collection which usually includes non-comparable pairs. Such non-comparable alternatives are often made comparable via a scalarization: the weighted sum method is one of the simplest but problems occur if the weights are not known (e.g., the `unknown weight scenario' in \cite[p. 72]{RoijersEtAl13}) or this method is not desirable at all. A new ranking (a.k.a., scalarization) method is proposed in this paper which takes into account a whole bunch of predefined linear scalarizations at once and is monotone with respect to a vector preoder generated by a convex cone. This method can also detect certain elements of the Pareto frontier which do not necessarily lie on the boundary of the convex hull of the set of alternatives.

The symbols $\N$ and $\R$ are used for the sets of natural (including 0) and real numbers, respectively.

Let $d \in \N\bs\{0,1\}$ and $C \subseteq \R^d$ be a proper closed convex cone. This means that $C$ is a closed set satisfying $sC = C$ for all $s \geq 0$, $C + C = C$, $C \not\in \{\varnothing, \R^d\}$. It generates a vector preorder $\leq_C$ (a reflexive and transitive relation which is compatible with addition and multiplication with non-negative numbers) via $y \leq_C z$ iff $z-y \in C$. Special cases are the zero cone $C = \{0\}$ and closed halfspaces $C = H^+(w) := \{z \in \R^d \mid w^T z \geq 0\}$ for $w \in \R^d\backslash\{0\}$ as well as, of course, $C = \R^d_+$ which generates the component-wise order. The symbol $y <_C z$ is used for $z - y \in \Int C$ where $\Int C \neq \varnothing$ is assumed.

Let $N \in \N\backslash\{0,1\}$ alternatives be given, i.e., a set $X = \{x^1, \ldots, x^N\} \subseteq \R^d$. The ranking problem consists in finding a function $r \colon X \to \R$ which ranks the alternatives in $X$. It is asked that such a ranking be compatible with $\leq_C$:
\begin{equation}
\label{EqRanking}
y \leq_C z \quad \Rightarrow \quad r(y) \leq r(z).
\end{equation}
A ranking function is called strict if $y <_C z$ implies $r(y) < r(z)$. Most difficulties in MCDM stem from the fact that the order relation $\leq_C$ is not total, i.e., there are non-comparable pairs: $y, z \in X$ with $y \not\leq_C z$, $z \not\leq_C y$. Of course, this is the case for the component-wise order generated by $C = \R^d_+$: one alternative can be better with respect to some criteria, but worse with respect to others. 

A higher value of the ranking functions is ``better" if the goal is to maximize with respect to the underlying vector order $\leq_C$; a lower value should be considered better if the goal is minimization. In this paper, ``better" is associated with a greater value of the rank function and therefore we also say that an alternative $z \in X$ dominates another $y \in X$ if $y \leq_C z$. Of course, this does not restrict generality.

The weighted sum scalarization is an easy way to find an $r$ satisfying \eqref{EqRanking}: take $w \in C^+ = \{v \in \R^d \mid \forall z \in C \colon v^\top z \geq 0\}$ ($C^+$ is called the dual of the cone $C$) and define $r(z) = w^T z$. This makes the ranking a very subjective procedure: changing the weight vector even slightly might result in a different ranking. Moreover, different decision makers (e.g., reviewers of a project) might have different weight vectors. Therefore, many other methods have been suggested such as TOPSIS, ELECTRE, AHP, PROMETHEE etc.

A ranking function always comes with a loss of information: pairs $y, z \in X$ which are not comparable w.r.t. $\leq_C$ are made comparable by assigning numbers $r(y), r(z)$ which can be compared  and the knowledge of the ranking numbers is not sufficient to decide if, say, $y$ is really better than $z$ or if they are not comparable with respect to $\leq_C$ and the better rank for $y$ is just the result of the ranking procedure itself. In our opinion, this issue is often neglected when discussing ranking methods, but essential for the new ranking method proposed below.

It originates from multivariate statistics where cone distribution functions were used to define quantiles of multi-dimensional random variables \cite{HamelKostner18}. It will be shown that these functions can be used to define rankings (this idea is due to \cite{Kostner20}) which enjoy very special rank reversal features. These features turn out to be useful properties instead of a nuisance since the situations in which a rank reversal occurs can be tracked and explained.

\section{Cone ranking functions}

The following definition introduces the basic concept of the paper. If $A$ is a finite set, $\#A$ denotes its cardinality, i.e., the number of elements in $A$.

\begin{definition}
\label{DefConeRanking} 
The functions $r_{X, w} \colon \R^d \to \N$ for $w \in C^+\backslash\{0\}$ and $r_{X, C} \colon \R^d \to \N$ defined by
\begin{align}
\label{EqW-Rank}
r_{X, w}(z) & = \#\{x \in X \mid x \in z - H^+(w)\} \quad \text{and} \\
\label{EqC-Rank}
r_{X, C}(z) & = \min_{w \in C^+} \#\{x \in X \mid x \in z - H^+(w)\} 
\end{align}
are called $w$-ranking and cone ranking function, respectively, for the set $X$.
\end{definition}

Clearly, $r_{X, C}(z) = \min_{w \in C^+} r_{X, w}(z)$. Moreover, $r_{X, w} = r_{X, H^+(w)}$. This function is the $N$th multiple of the empirical lower cone distribution functions from \cite{HamelKostner18, HamelKostner22} in which $X$ is considered as a set of multidimensional data points, or, more general, the values of a random vector. Note that $r_{X, w}$ and $r_{X, C}$ are defined for all $z \in \R^d$ which means that every point $z \in \R^d$ can be ranked with respect to the set $X$ of given alternatives.

\begin{proposition}
The functions $r_w$ and $r_C$ are strict ranking functions. Moreover, one has
\begin{equation}
\label{EqAffEqui}
\forall z \in \R^d \colon r_{AX + b, AC}(Az + b) = r_{X, C}(z)
\end{equation}
for an invertible matrix $A \in \R^{d \times d}$ and a vector $b \in \R^d$ where $AX+b = \{Ax^1+b, \ldots, Ax^N+b\}$. Finally, if $D \subseteq \R^d$ is another closed convex cone with $C \subseteq D$, then $r_{X, C}(z) \leq r_{X, D}(z)$ for all $z \in \R^d$; in particular, $r_{X, C}(z) \leq r_{X, H^+(w)}(z)$ for all $w \in C^+$ and all $z \in \R^d$.
\end{proposition}

{\sc Proof.} Assume $y \leq_C z$. Then $w^T y \leq w^\top z$ for all $w \in C^+$. Hence $x \in y - H^+(w)$ implies $x \in z - H^+(w)$. This implies $r_{X, w}(y) \leq r_{X, w}(z)$ for all $w \in C^+$ and consequently $r_{X, C}(y) \leq r_{X, C}(z)$. If $y <_C z$, then $y \in z - \Int H^+(w)$  for all $w \in C^+\backslash\{0\}$, hence $z \not\in y - H^+(w)$. On the other hand $x \in y - H^+(w) \subseteq z - H^+(w)$. Hence $r_{X, w}(y) + 1 \leq r_{X, w}(z)$ for all $w \in C^+\backslash\{0\}$ which gives the result. The straightforward proof of the affine equivariance property \eqref{EqAffEqui} can be found in \cite{HamelKostner18}. The last claim follows from the definition of the ranking functions $r_{X, C}$, $r_{X, D}$ together with a well-known property of dual cones: if $C \subseteq D$, then $D^+ \subseteq C^+$.

\smallskip
{\bf Interpretation.} The number $r_w(z)$ is the number of alternatives which have a lower or the same weighted sum with weight distribution $w$ than $z$; the minimum over these numbers over all reasonable weight vectors gives $r_{X, C}(z)$. This means that no matter which feasible weight vector is chosen, $z$ has a greater weighted sum than at least $r_{X, C}(z)$ alternatives. If the task is to find an alternative which is maximal with respect to $\leq_C$, then one would look for $x \in X$ with $r_{X, C}(x)$ as large as possible--$x$ dominates as many points as possible no matter which weighted sum is chosen. This makes it clear that a ranking via $r_w$ or $r_C$ is a relative one: there is no objective scale, the alternatives are only mutually compared, not with respect to an outside scale (such as temperature, for instance). Roughly speaking, the function $r_{X, C}$ ranks an alternative higher if the minimal number of alternatives it dominates with respect to a feasible weight distribution is greater. Such a ranking is desirable for example in cases of indices for countries, economies, projects, candidates etc. where choices can be made only among a pool of available alternatives and one wishes to select the relative best. 

\smallskip
{\bf Potential outcomes.} Points which are not comparable with respect to $\leq_C$ can have the same or (even very) different values of $r_{X, C}$. As an example, consider the set $X$ of black and yellow dots in Figure \ref{fig:RankRev1} below with cone $C = \R^2_+$: the upper right black dot in the $2^{\text{nd}}$ quadrant has rank 2 while the black dot at the intersection of the dotted lines has rank 6. Of course, the two points are not comparable. Thus, a low ranking can have two different reasons: first, the alternative in question is rarely comparable to other alternatives, secondly it is dominated by many other alternatives.

One might call an alternative an outlier if it is not comparable to (and thus not dominated by) many others and has a very low $r_{X, C}$-value compared to the best ranked alternative--it has very different (not necessarily worse) features than the rest. It could be useful to identify such alternatives, e.g., for recommender systems since it could make sense to mix in such an alternative sometimes as a recommendation to provide options outside the usual ``bubble."

\smallskip
{\bf Pareto optimality.} A point $\bar x \in X$ is (Pareto) maximal in $X$ with respect to $\leq_C$ iff $x \in X$ and $\bar x \leq_C x$ imply $x \leq_C \bar x$ (there is no "strictly greater" alternative). If $C$ is pointed, i.e., $C \cap (-C) = \{0\}$, then $\leq_C$ is antisymmetric and $x \in X$, $\bar x \leq_C x$ implies $x = \bar x$. The following result provides a sufficient condition for maximality in terms of the ranking function $r_{X, C}$.

\begin{theorem}
\label{ThmMaximality}
Let $C$ be a pointed polyhedral cone, i.e., it is pointed and the intersection of a finite number of halfspaces. If $r_{X, C}(\bar x) = \max_{x \in X} r_{X, C}(x)$, then $\bar x \in X$ is maximal. The converse is not true in general.
\end{theorem}

{\sc Proof.} Assume there is a point $y \in X$, $y \neq \bar x$ with $\bar x \leq_C y$. Without loss of generality one can assume that $y$ is maximal in $X$; otherwise it can be replaced by a maximal one since there are only finitely many alternatives. Since $\bar x$ has maximal ranking and $\bar x \leq_C y$, one has $r_{X, C}(\bar x) = r_{X, C}(y)$. Assume $\bar w \in C^+$ provides the minimum in \eqref{EqC-Rank} for $\bar x$ and $v \in C^+$ for $y$: they exist since $X$ is finite and $C$ is polyhedral. Assume first $\bar w^\top \bar x < \bar w^\top y$. Then one has $r_{X, \bar w}(\bar x) = r_{X, C}(\bar x) = r_{X, C}(y) = r_{X, v}(y)$. Since $y - x \in C$ and $v \in C^+$ one also has $v^\top \bar x \leq v^\top y$.

In view of \eqref{EqW-Rank}, denote $X^\leq(w, x) := \{z \in X \mid z \in x - H^+(w)\} = \{z \in X \mid w^\top z \leq w^\top x\}$ for $w \in C^+$, $x \in X$. Then $\#X^\leq(\bar w, \bar x) = \#X^\leq(v, y)$ since $r_{X, \bar w}(\bar x) = r_{X, v}(y)$. Further, $\#X^\leq(v, \bar x) \leq \#X^\leq(v, y)$ since $v^\top \bar x \leq v^\top y$ and $\#X^\leq(\bar w, \bar x) \leq \#X^\leq(v, \bar x)$ because of the minimality property of $\bar w$. Altogether, $\#X^\leq(\bar w, \bar x) = \#X^\leq(v, y) \leq \#X^\leq(v, \bar x) \leq \#X^\leq(v, y)$, hence equality holds for all of these numbers. Thus $v$ provides the minimum in \eqref{EqC-Rank} for $r_{X, C}(\bar x)$ as well as for $r_{X, C}(y)$. Hence, without loss of generality, one can replace $\bar w$ by $v$. Relabeling $v$ by $\bar w$ one gets $r_{X, \bar w}(\bar x) = r_{X, C}(\bar x) = r_{X, C}(y) = r_{X, \bar w}(y)$ and $\bar w^\top \bar x = \bar w^\top y$.

Assuming this, define the polytope $P_y(\bar x)$ as the convex hull of $X^\leq(\bar w, \bar x) \backslash \{y\}$. The set $P_y(\bar x) - C$ includes $\bar x$, but not $y$ since $y$ is maximal and $C$ is pointed. Separating one gets $w \in C^+$ with $\max_{x \in P_y(\bar x)} w^\top x < w^\top y$. In particular, $w^\top \bar x < w^\top y$. Define $w(s) = \bar w + sw \in C^+$. Then one has $y \in X^\leq(\bar w, \bar x)$ but $y \not\in X^\leq(w(s), \bar x)$ for $s > 0$. Since $X$ is a finite set, there is $s> 0$ small enough such that $X^\leq(w(s), \bar x) \subsetneq X^\leq(\bar w, \bar x)$. This contradicts $r_{X, C}(\bar x) = r_{X, \bar w}(\bar x)$ which concludes the proof.

\medskip
\begin{figure}[h]
	\centering
	\begin{minipage}{.6\textwidth}
	\centering
	\includegraphics[width=0.8\textwidth]{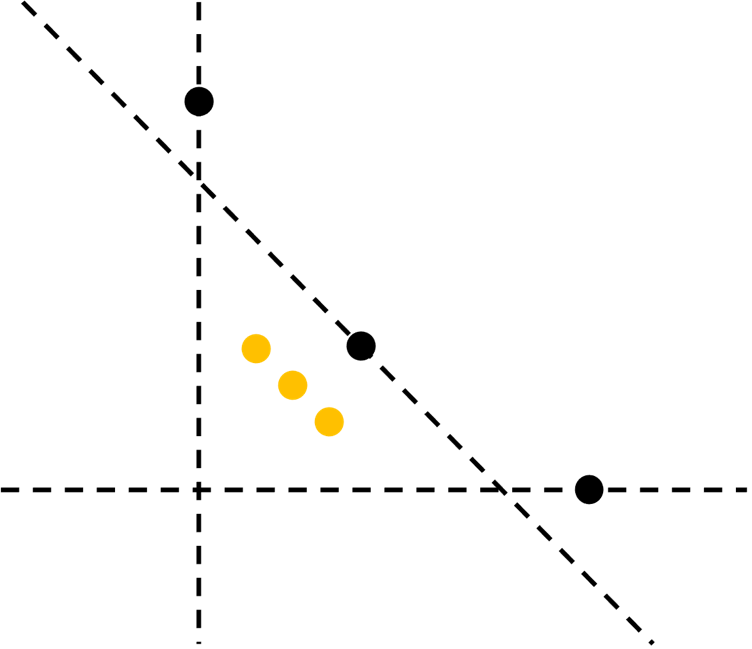}
	\caption{Non-convex Pareto frontier}
	\label{fig:Concave}
	\end{minipage}
\end{figure}

Figure \ref{fig:Concave} shows a situation where points with a maximal value of $r_{X, C}$ are not necessarily on the ``convex part" of the Pareto frontier, i.e., on the Pareto frontier of the convex hull of the points representing the alternatives. First, consider only the three black points: they all have rank 1 and are Pareto maximal with respect to the order generated by $C = \R^2_+$. If one adds the three yellow points, then the three black ones are still Pareto maximal and the one in the middle gets the maximal ranking which now is 4. Clearly, this point does not belong to the convex hull of the now 6 data points. By the way of conclusion, this example shows that the ranking function $r_{X, C}$ can detect maximal alternatives not belonging to the ``convex part" of the Pareto frontier--even though only linear scalarizations enter the definition in \eqref{EqC-Rank}. On the other hand, it also shows that the rank of a point particularly depends on how many other points it dominats with respect to the order generated by the cone: if one placed the three yellow points in Figure \ref{fig:Concave} close to the upper left or the lower right black point, one could generate maximal ranking for either of them. This feature will be further discussed in the sequel.

\section{Rank reversal for cone ranking functions}

Rank reversal of one or another type occur for virtually every MCDM method, and its implications are still highly debated. One may compare, for example, \cite{AiresFerreira18, WangLuo09} for surveys and \cite{GarciaCascalesLamata12} for TOPSIS, \cite{Maleki13, MajumdarEtAl21} for AHP, \cite{WangTriantaphyllou08, FigueiraRoy09} for ELECTRE and \cite{MareschalDeSmetNemery08, VerlyDeSmet13, DejaegereDeSmet22} for PROMETHEE.

Let $z^1, \ldots, z^M \in \R^d$ be alternatives which are added to $X$ and denote $Z = X \cup\{z^1, \ldots, z^M\}$. A rank reversal occurs if
\begin{equation}
\label{EqRankReversal}
r_{X, C}(x) < r_{X, C}(y) \quad \text{and} \quad r_{Z, C}(y) < r_{Z, C}(x)
\end{equation}
for $x, y \in X$. A weak rank reversal occurs if (only) one of the two inequalities in \eqref{EqRankReversal} is replaced by $\leq$. Clearly, a rank reversal is not possible if $x \leq_C y$ by definition of $r_{X, C}, r_{Z, C}$.

\smallskip
{\bf Result 1.} If $x$ and $y$ are not comparable with respect to $\leq_C$ and $r_{X, C}(x) \leq r_{X, C}(y)$, then one can add $M := r_{X, C}(y) -  r_{X, C}(x) + K$ alternatives $z^1, \ldots, z^M$ and get  $r_{X, C}(x) = r_{X, C}(y) + K$, i.e., a (weak) rank reversal occurs. Indeed, if $x$ and $y$ are not comparable, one can add $M$ points which are also not comparable to $y$, but dominated by $x$, i.e., $z^m \leq_C x$ for $m = 1, \ldots, M$. 

\begin{figure}[h]
	\centering
	\begin{minipage}{.4\textwidth}
	\centering
	\includegraphics[width=0.8\textwidth]{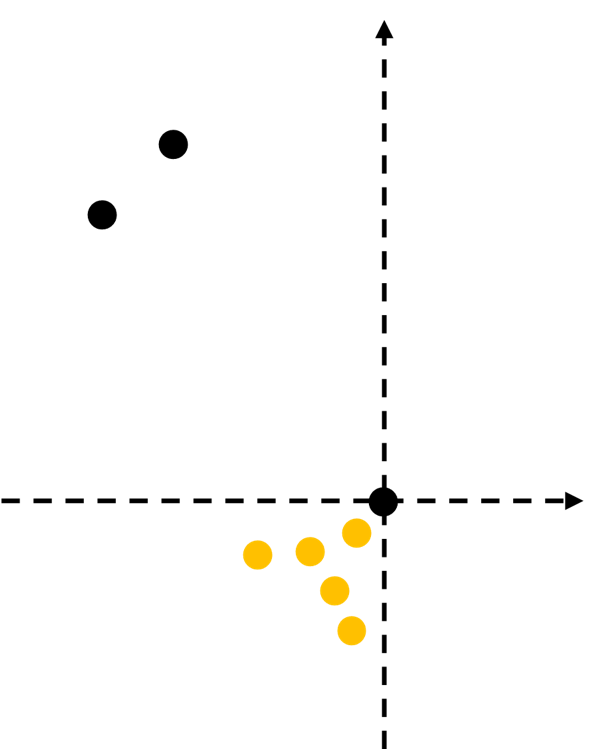}
	\caption{Rank reversal 1}
	\label{fig:RankRev1}
	\end{minipage}
\end{figure}

\smallskip
Figure \ref{fig:RankRev1} shows an example: the rank of the lower black point jumps from 1 to 6 if the 5 yellow points are added while the rank of the upper right black point remains 2.

Understanding this rank reversal feature contributes to understanding the proposed ranking method. First, a high ranking $r_{X, C}(y)$ compared to $r_{X, C}(x)$ does not mean that alternative $y$ is necessarily much better than $x$. It only means that $y$ dominates a greater number of alternatives if the worst weight distributions are chosen, i.e., the $w$'s which provides the minimum in \eqref{EqC-Rank} for $x$ and $y$, respectively. Still, $x$ and $y$ can be non-comparable with respect to the original order $\leq_C$. In such a case, this could indicate that $x$ has very different features compared to $y$ and all alternatives dominated by $y$ (the yellow points in Figure \ref{fig:RankRev1}). One might say that in such a case $y$ is a ``common" and $x$ is a ``rare" alternative and one would need to decide between a safe (common) and an adventureous (rare) option which are not comparable. To test the occurence of this feature, one can rank the alternatives according to $r_{X, C}$, then remove all alternatives which are dominated by the best one(s) and repeat the ranking with the reduced set. In the example of Figure \ref{fig:RankRev1}, one would remove the yellow points after ranking all yellow/black points and get the upper right black point as the ``rare" option. Note, however, that very different ranks $r_{X, C}(x)$, $r_{X, C}(x)$ can also occur in cases where $x, y \in X$ dominate the same number of other alternatives.

\smallskip
{\bf Result 2.} If one adds one alternative $z:=z^1$ to $X$, i.e., $Z = X \cup\{z\}$, then $r_{Z, C}(x) = r_{X, C}(x)$ for $x \in X$ or $r_{Z, C}(x) = r_{X, C}(x) + 1$. In the second case, weak rank reversal may occur. 

\begin{figure}[h]
	\centering
	\begin{minipage}{.5\textwidth}
	\centering
	\includegraphics[width=0.9\textwidth]{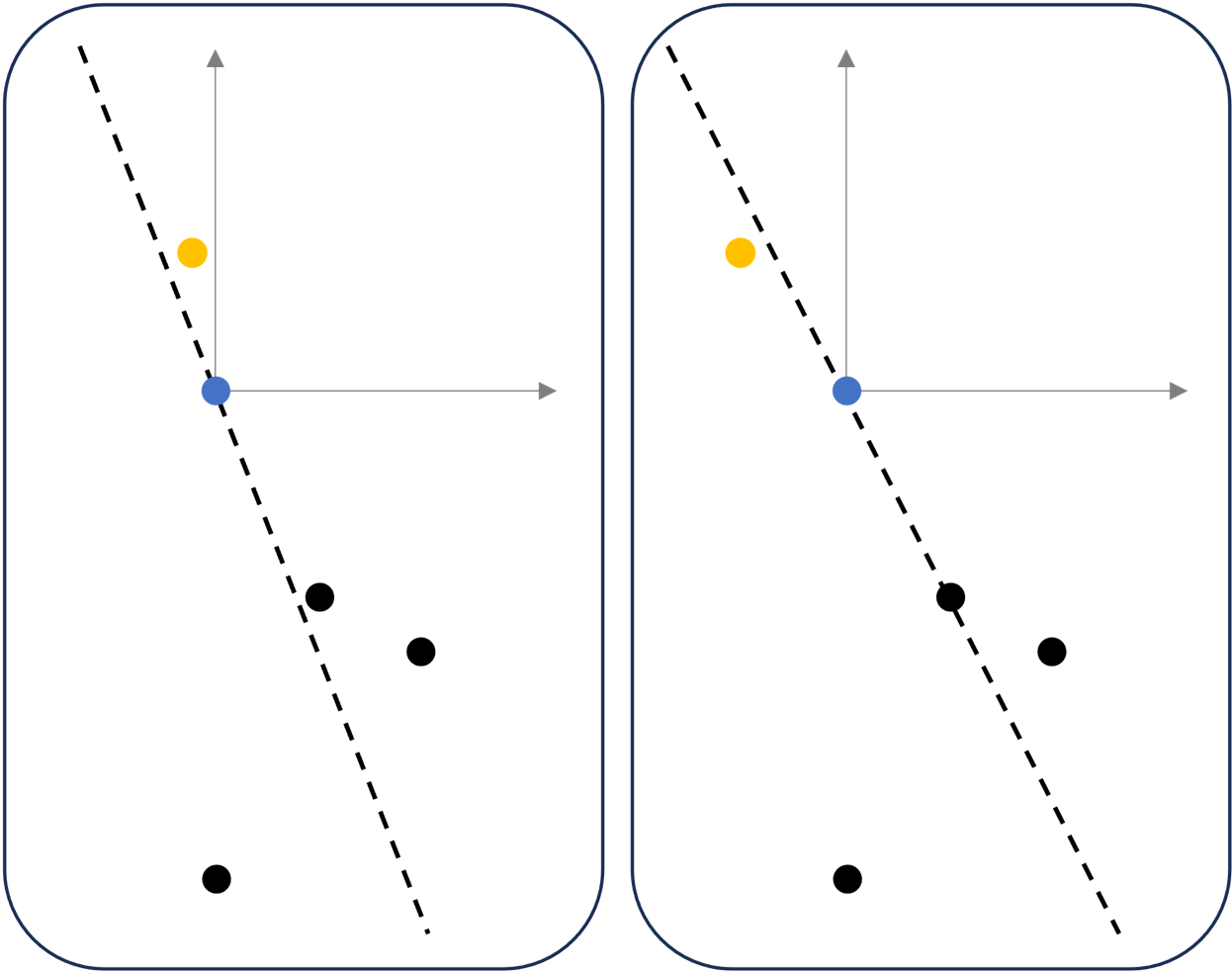}
	\caption{Rank reversal 2}
	\label{fig:RankRev2}
	\end{minipage}
\end{figure}

This is illustrated by the example shown in Figure \ref{fig:RankRev2}. The set $X$ comprises the blue point $x$ as well as the black ones, and one has $r_{X, C}(x)=2$ (note the doted line in the left picture). It depends on the location of the added yellow point $z$ if the rank of $x$ changes: in the left picture it does not change, in the right picture one has $r_{Z, C}(x) = 3$. The ranks of the black points do not change in either situation, they are 1, 2 and 2, respectively.  
Thus, the (new) rank $r_{Z, C}(x)$ does not only depend on $x$ and $Z$, but also on the location of the other alternatives.

\smallskip
{\bf Result 3.} If $X = \{x, y\}$ has only two points, then 3 cases are possible. Case 1: $x$ and $y$ are not comparable w.r.t. $\leq_C$ and hence $r_{X, C}(x) = r_{X, C}(y) = 1$. Case 2: one has, without loss of generality, $x \leq_C y$, $y \not\leq_C x$ and hence $r_{X, C}(x) = 1$, $r_{X, C}(y) = 2$. Case 3: $x \leq_C y$, $y \leq_C x$ and hence  $r_{X, C}(x) = r_{X, C}(y) = 2$. 

In the first case, a weak rank reversal may occur if alternatives are added, but this is not possible in the second case: $y$ will always be higher ranked than $x$. In the third case, (only) a weak rank reversal is possible. This shows that intransitivity effects cannot occur which is due to \eqref{EqRanking} and the transitivity of $\leq_C$. The third case, however, cannot occur if $\leq_C$ is antisymmetric, i.e., precisely if $C \cap (-C) = \{0\}$.

\begin{figure}[h]
	\centering
	\begin{minipage}{.5\textwidth}
	\centering
	\includegraphics[width=0.9\textwidth]{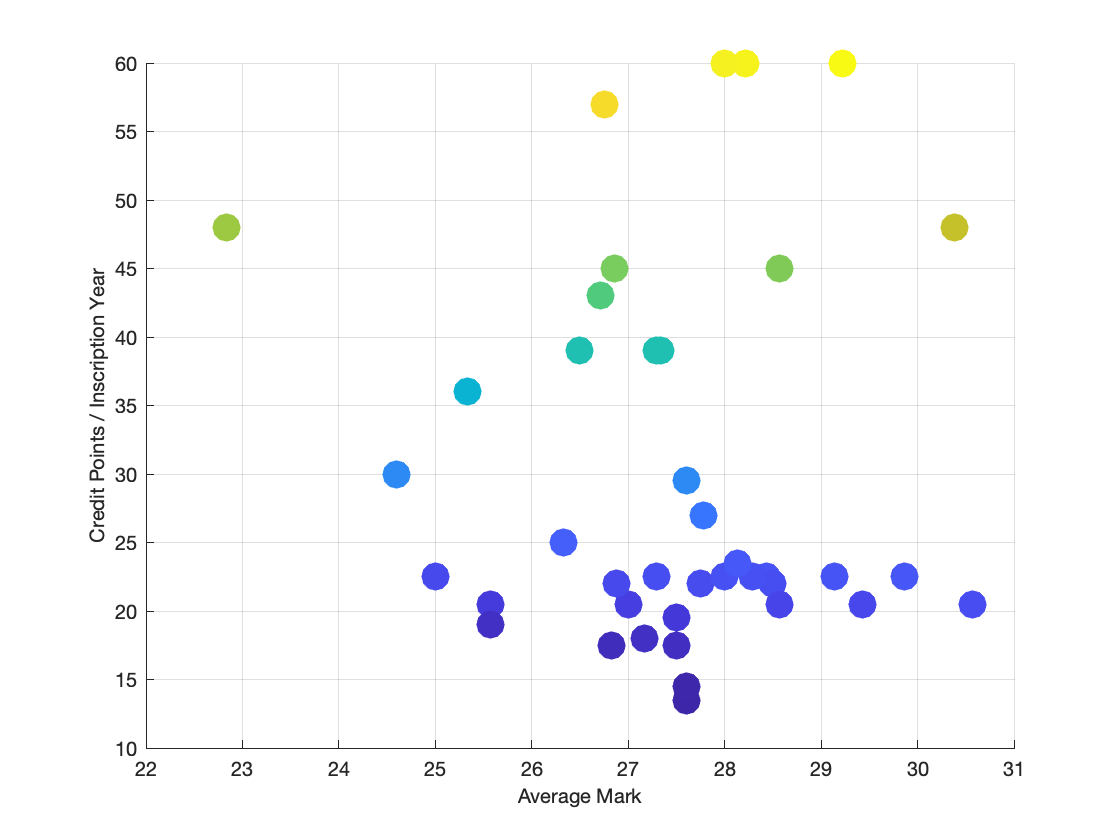}
	\caption{Student ranking with TOPSIS}
	\label{fig:StudentTopsis}
	\end{minipage}
\end{figure}

\begin{figure}[h]
	\centering
	\begin{minipage}{.5\textwidth}
	\centering
	\includegraphics[width=0.9\textwidth]{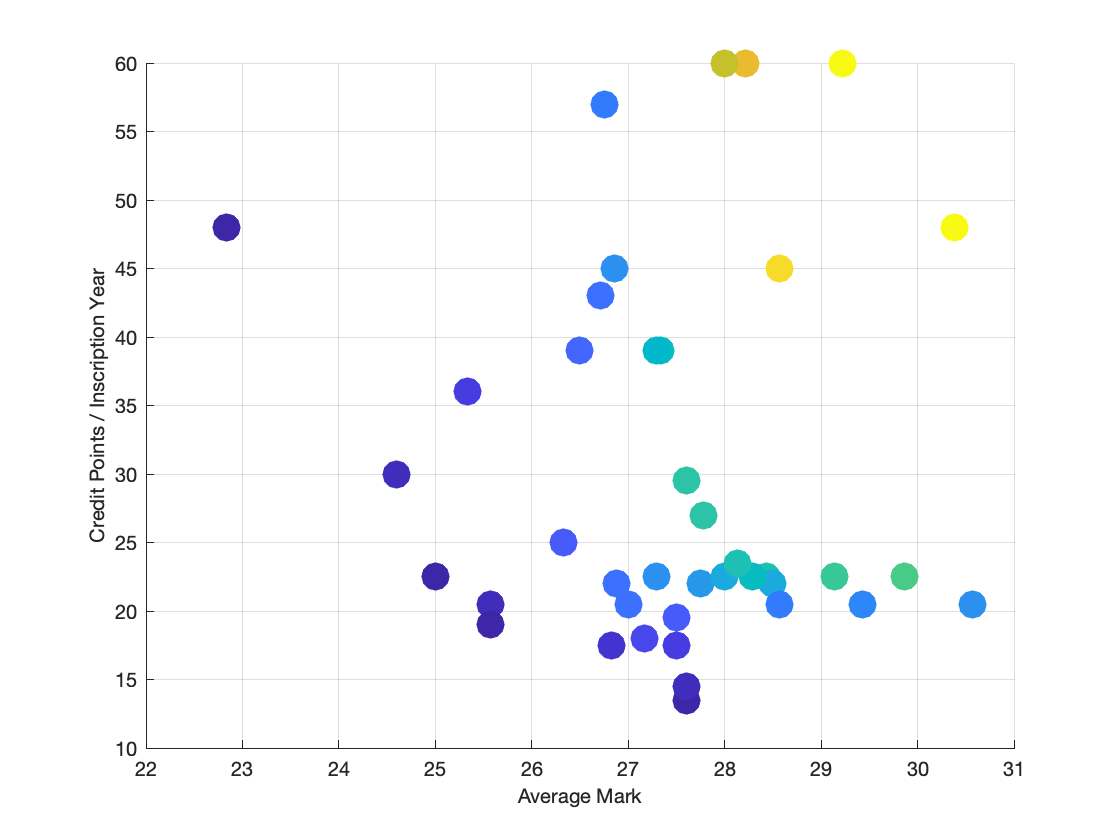}
	\caption{Student ranking with $r_{X,\R^2_+}$}
	\label{fig:StudentCone}
	\end{minipage}
\end{figure}

\medskip
The cone ranking function can produce results which are quite different from those generated by standard MCDM tools such as TOPSIS. The ranking of a student cohort is discussed as an illustrating example. The two criteria ``average mark in exams" and ``credit points achieved in a given time interval" are used. In the pictures, higher ranked individuals appear in lighter colours. Figure \ref{fig:StudentTopsis} shows the ranking obtained with TOPSIS: this method puts more weight--a little counterintuitively--to the credit point criterion. Figure \ref{fig:StudentCone} shows the result according to the cone ranking function with $C = \R^2_+$: it gives higher rankings to alternatives in the upper right area.

 A similar example has been discussed in \cite[Example 6.3]{HamelKostner22} from a statistical point of view.

\section{Discussion of the cone}

In the examples above, the ``MCDM cone" $C = \R^d_+$ was used. However, different cones are sometimes of advantage. The elements of the dual cone $C^+$ in Definition \ref{DefConeRanking} can be seen as potential weight vectors for the $d$ different criteria. One may also observe that the condition $x \in z - H^+(w)$ in  \eqref{EqW-Rank}, \eqref{EqC-Rank} means $w^T(x - z) \leq 0$, thus it is positively homogeneous in $w$. Therefore, one can restrict the set of $w$'s to $B^+ = \{w \in \R^d_+ \mid w_1 + \ldots + w_d = 1\}$ if $C = \R^d_+$. The set $B^+$ includes all potential weight vectors for the $d$ criteria. If the decision maker wants to make sure that each criterion is given a minimal and a maximal weight, say $0 \leq w_i^{min} \leq 1$ and $0 \leq w_i^{max} \leq 1$ for $i \in \{1, \ldots, d\}$, respectively, then one can consider the set
\[
W = \{w \in B^+ \mid w_i^{min} \leq w_i \leq w_i^{max}, \, i \in \{1, \ldots, d\} \}
\]
and replace $C^+$ in \eqref{EqW-Rank}, \eqref{EqC-Rank} by $W$.

For example, one may assign a minimal and a maximal weight to the average mark (and/or to the number of credit points achieved) for the student ranking, e.g., as the result of a discussion in an evaluating panel. Let say, these numbers are $w_1^{min}, w_1^{max} \in (0, 1)$. Then one defines $W = \{w \in \R^2_+ \mid w_1 + w_2 = 1, w_1^{min} \leq w_1 \leq w_1^{max}\}$ and $C^+ = \{sw \mid w \in W, s \geq 0\}$ and $C = (C^+)^+ = \{z \in \R^2 \mid \forall w \in C^+ \colon w^T z \geq 0\}$. The cone $C^+$ now is smaller then $\R^2_+$, the cone $C$ bigger than $\R^2_+$ according to the relationships for dual cones. This is also the underlying idea for panel/multi-judge MCDM in \cite{Kostner20} and motivates the use of a general cone $C$ instead of just $\R^d_+$.

Moreover, this procedures gives more flexibility to the decision maker since the weight distribution does not have to be fixed in advance: the cone ranking function is a worst case ranking with respect to a variety of weight distributions. It is also useful to consider the minimizer(s) in \eqref{EqC-Rank}: these are the weight vectors which gives the worst $w$-ranking of an alternative. Thus, if the decision maker has a preferred weight distribution (very) different from the minimizers in $r_{X, C}(x)$ for $x \in X$, then the corresponding $w$-ranking of $x$ might even be much better than its $C$-ranking. 

\begin{figure}[h]
	\centering
	\begin{minipage}{.5\textwidth}
	\centering
	\includegraphics[width=1\textwidth]{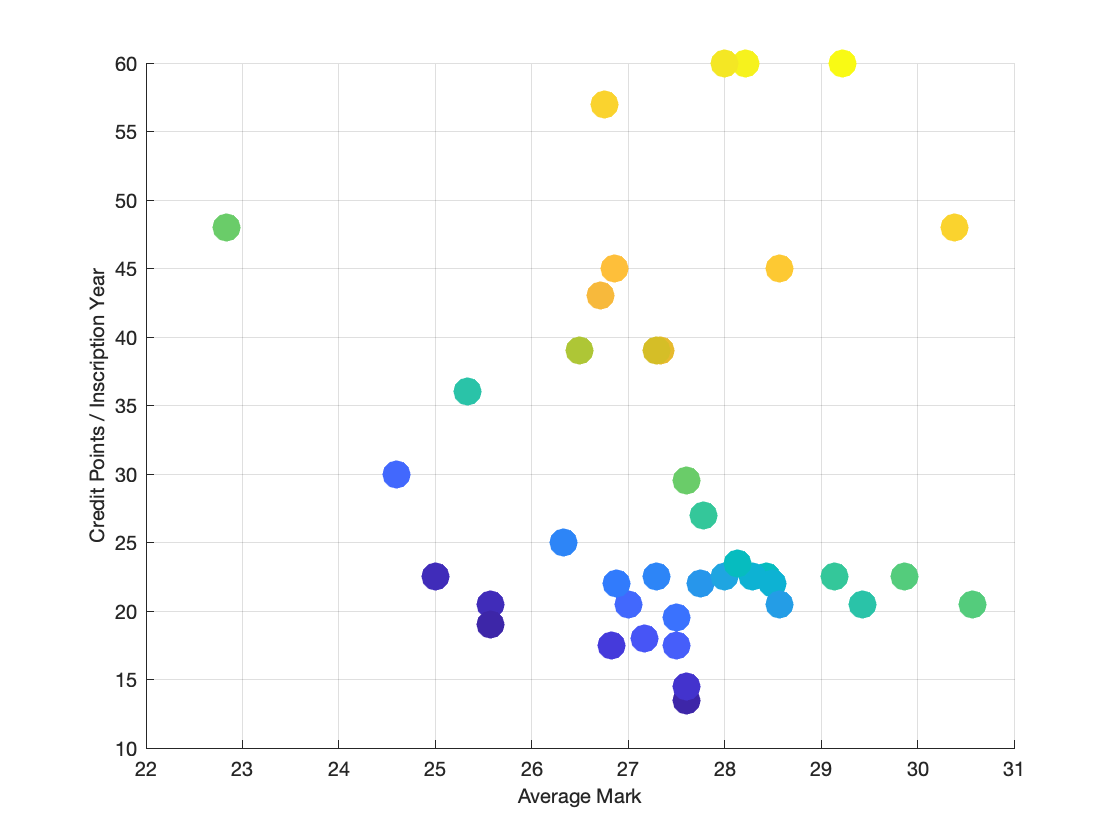}
	\caption{Modified student ranking}
	\label{fig:StudentBiggerCone}
	\end{minipage}
\end{figure}

In Figure \ref{fig:StudentBiggerCone}, the ranking of the same student cohort is depicted where the cone $C$ is generated by the two vectors $(0.7, 0.3)$, $(0.8, 0.2)$. This means that the minimal weight assigned to average grade is $70\%$, the maximal weight is $80\%$. One may observe that this creates higher ranking in the upper left as well as lower right part of the point cloud due to the fact that one now has lower as well as upper bounds for the weights which are strictly less than $100\%$. In particular, the remote point on the upper left with a considerable low average mark now has a better ranking.

The dual cone now is smaller which means that the original cone $C$ is bigger as can be seen in Figure \ref{fig:ModifiedCone} below: this creates compensation opportunities, i.e., students can compensate for a low average mark with a higher number of credit points within the same time interval or vice versa. This shows that the ranking of such alternatives (students, candidates, projects etc.) requires to answer the question of how much surplus in one attribute can compensate a given deficit in another attribute. 

\begin{figure}[h]
	\centering
	\begin{minipage}{.4\textwidth}
	\centering
	\includegraphics[width=0.9\textwidth]{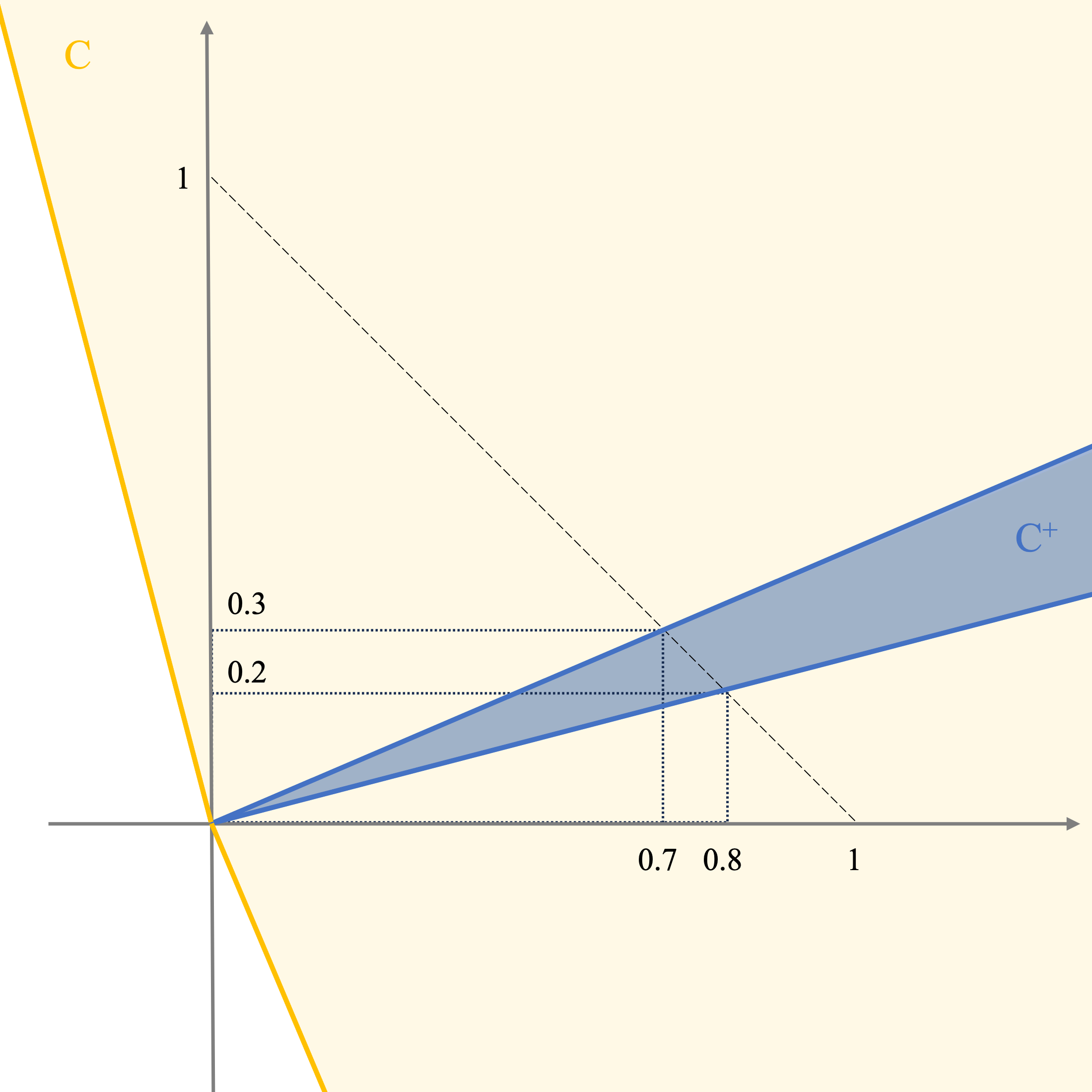}
	\caption{Modified cone}
	\label{fig:ModifiedCone}
	\end{minipage}
\end{figure}

\section{Extension to rankings of sets}

The comparison and ranking of sets as a problem was introduced to MCDM by Zitzler and Thiele in the highly cited paper \cite{ZitzlerThiele99}. It was motivated by the need to compare the output sets of different MCDM algorithms which include potentially non-dominated alternatives. In the follow-ups \cite{ZitzlerEtAl03, ZitzlerThieleBader10}, several order relations for sets can be found, especially the "dominates" relation (along with strict and weak versions) which coincides with the following order relation for sets: a set $A \subseteq \R^d$ dominates another $B \subseteq \R^d$, written $A \succeq_C B$, if
\[
\forall b \in B, \; \exists a \in A \colon a \geq_C b.
\]
In the vector preorder setting, this is equivalent to $A - C \supseteq B$. This relation is one of two which 
form the building block for the theory of set optimization: compare \cite{HamelEtAl15Incoll} for surveys of the relevant literature. It has been known before, for example, in Economics: compare \cite{HamelLoehne20MMOR-Pre} for some historic comments. To the best of our knowledge, the Thiele-Zitzler approach was developed independently of the set optimization literature (and vice versa) which can also be concluded from \cite{ZitzlerThieleBader10}: in this reference, the twin relation to $\succeq_C$ can be found (compare \cite[p. 61, Sect. C]{ZitzlerThieleBader10}) which is more suited for minimization.

The performance measure $\mathcal C$ in \cite[Sect. III, (7)]{ZitzlerThiele99} gives the value $\mathcal C(A, B) = 1$ precisely if $A \succeq_C B$ for $A, B \subseteq X$ since it is the number of alternatives in $B$ which are dominated by at least one alternative in $A$ relative to the number of alternatives in $B$.

On the other hand, set extensions of the lower cone distribution function from \cite{HamelKostner18} were shown to be in the same relation with set-valued quantiles as the cumulative distribution function with quantile function for univariate random variables--they form a Galois connection (and are even inverse to each other if the underlying random variable is continuous). These set extensions are monotone with respect to the underlying set order relation. A parallel extension of the ranking functions from Definition \ref{DefConeRanking} can be used to rank sets of alternatives instead of single alternatives.

In the following, the symbols $\mathcal P(X)$, $\mathcal P(\R^d)$ denote the power sets of $X$ and $\R^d$, respectively--including the empty set in each case.

\begin{definition}
\label{DefSetRanking} The functions $R^\trido_w, R^\trido_C \colon \mathcal P(\R^d) \to \N$ defined by
\[
R^\trido_w(A) = \sup_{a \in A} r_{X, w}(a) \quad \text{and} \quad R^\trido_C(A) = \sup_{a \in A} r_{X, C}(a) 
\]
are called set $w$-ranking for $w \in C^+$ and set cone ranking function, respectively, generated by the set $X$.
\end{definition}

Clearly, if $A$ is a finite set--especially if $A \in \mathcal \mathcal P(X)$--then the suprema in Definition \ref{DefSetRanking} can be replaced by maxima. Moreover, one of course has $R^\trido_w(\{a\}) = r_{X, w}(a)$ for singleton sets $A = \{a\}$.

\begin{proposition}
\label{PropSetRankMonotonicity}
Both $R^\trido_w$ and $R^\trido_C$ are monotone with respect to $\succeq_C$. 
\end{proposition}

{\sc Proof.} First, we show $R^\trido_C(A - C) = R^\trido_C(A)$. Indeed, one has $a - z \leq_C a$ for all $a \in A$, $z \in C$, hence
\[
R^\trido_C(A - C) = \sup_{a \in A, z \in C} r_{X, C}(a - z) \leq \sup_{a \in A} r_{X, C}(a) = R^\trido_C(A)
\]
by monotonicity of $r_{X, C}$. On the other hand, $A - C \supseteq A$ implies
\[
R^\trido_C(A - C) = \sup_{y \in A - C} r_{X, C}(y) \geq \sup_{a \in A} r_{X, C}(a) = R^\trido_C(A),
\]
hence equality holds true. Next, take $A, B \subseteq \R^d$ with $B \subseteq A - C$. Then
\[
R^\trido_C(B) \leq R^\trido_C(A - C) = R^\trido_C(A)
\]
which proves the claim. The special case $C = H^+(w)$ yields the same result for $R^\triup_w$. 

\medskip
{\bf Interpretation.} The definition of $R^\trido_w$, namely $R^\trido_w(A) = \sup_{a \in A}  \#\{x \in X \mid w^\top x \leq w^\top a\}$, shows that $R^\trido_w(A)$ asks to look for $a \in A$ which gives the maximal number of points in $X$ which have a weighted sum less than the one of $a$ with weight vector $w$. Consequently, $R^\trido_C(A) = \sup_{a \in A} \min_{w \in C^+} \#\{x \in X \mid x \in a - H^+(w)\}$ ask for elements $a \in A$ for which the minimal number of points in $X$ with weighted sum less than the one for $a$ is maximal. Thus, it can be understood as a max-min performance measure for sets $A$: selecting a set is done by looking for best alternatives in sets with respect to $r_{X, C}$.

The comparison with the performance measure $\mathcal C$ from \cite[Sect. III, (7)]{ZitzlerThiele99} cannot be made directly since the latter is a function of two sets. However, if one chooses the set $X$ of all $N$ alternatives as one, then the function on $\mathcal P(\R^d)$ defined by
\[
\mathcal C_X(A) = N \cdot \mathcal C(A, X) = \#\cb{x \in X \mid \exists a \in A \colon a \geq_C x}
\]
is monotone with respect to $\succeq_C$ as one may easily check. This raises the question what the relationship between $R^\trido_C$ and $\mathcal C_X$ is.

If $A$ includes all non-dominated elements of $X$, then $\mathcal C_X(A) = N$ since $X$ has only finitely many elements and thus enjoys the upper domination property, i.e., each element of $X$ is dominated by a non-dominated element. Moreover, $\mathcal C_X(X) = N$. Therefore, $\mathcal C_X(A)/N = \mathcal C(A, X)$ for $A \subseteq X$ can be understood as a measure for the performance of $A$ relative to $X$.

\medskip
{\bf Potential outcomes.} First, one always has $R^\trido_C(\{z\}) \geq \mathcal C_X(\{z\})$ for all singleton sets with $z \in \R^d$. This is no longer true for non-singleton sets (compare the following examples).

\begin{figure}[h]
	\centering
	\begin{minipage}{.99\textwidth}
	\centering
	\includegraphics[width=\textwidth]{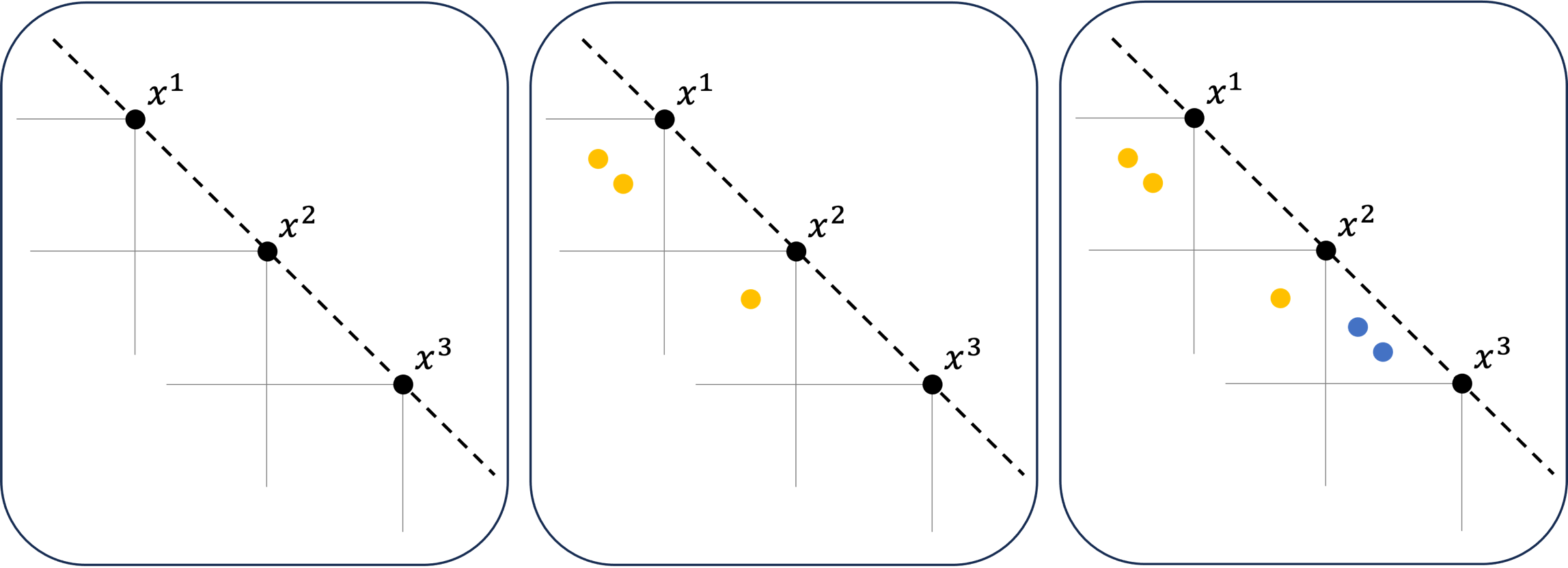}
	\caption{Set ranking}
	\label{fig:SetRanking}
	\end{minipage}
\end{figure}

Moreover, the ranking of sets via $R^\trido_C$ can be different from the one via $\mathcal C_X$. Figure \ref{fig:SetRanking} shows three examples where in all cases the sets $A^i = \{x^1, x^2, x^3\}\bs\{x^i\}$, $i = 1, 2, 3$, are compared. Depending which points are added to the set $X$ the rankings of $A^1, A^2, A^3$ change.

First, the left picture in Figure \ref{fig:SetRanking} with no additional points gives $\mathcal C_X(A^i) = 2$ for $i = 1, 2, 3$, i.e., $\mathcal C_X$ ranks all three sets as equal. On the other hand, $r_{X, C}(x^1) = 1$, $r_{X, C}(x^2) = 2$, $r_{X, C}(x^3) = 1$ and thus $R^\trido_C(A^1) = \max\{2, 1\} = 2$, $R^\trido_C(A^2) = \max\{1, 1\} = 1$, $R^\trido_C(A^3) = \max\{1, 2\} = 2$ which means that $A^2$ is ranked lower than the sets $A^1$, $A^3$ which are ranked equal. Since in all cases the cone is $C = \R^2_+$ one might say that $R^\trido_C$ gives a ranking which ranks sets higher if they include ``compromise" non-dominated points such as $x^2$ in this example.

Secondly, the center picture of Figure \ref{fig:SetRanking} shows a situation where three dominated points are added which are not comparable with each other. This changes the $\mathcal C_X$-ranking to $\mathcal C_X(A^1) = 3$, $\mathcal C_X(A^2) = 4$ and $\mathcal C_X(A^3) = 5$. On the other hand, one has $r_{X, C}(x^1) = 3$, $r_{X, C}(x^2) = 3$, $r_{X, C}(x^3) = 1$ and thus $R^\trido_C(A^1) = \max\{3, 1\} = 3$, $R^\trido_C(A^2) = \max\{3, 1\} = 3$, $R^\trido_C(A^3) = \max\{3, 3\} = 3$. The $R^\trido_C$-ranking is the same for all three sets while the $\mathcal C_X$-ranking is different.

Thirdly, in the right picture of Figure \ref{fig:SetRanking} two more points are added which are not comparable to any of the points in the previous situation. This does not change the $\mathcal C_X$-ranking compared to the center picture. On the other hand, $r_{X, C}(x^1) = 3$, $r_{X, C}(x^2) = 5$, $r_{X, C}(x^3) = 1$ and thus $R^\trido_C(A^1) = \max\{5, 1\} = 5$, $R^\trido_C(A^2) = \max\{3, 1\} = 3$, $R^\trido_C(A^3) = \max\{3, 5\} = 5$ which gives a different $R^\trido_C$-ranking compared to the previous case. 

In particular, these example also show that a set can be ranked higher with respect to the $R^\trido_C$-ranking and lower with respect to the $\mathcal C_X$-ranking as well as vice versa.

\begin{figure}[h]
	\centering
	\begin{minipage}{.5\textwidth}
	\centering
	\includegraphics[width=.8\textwidth]{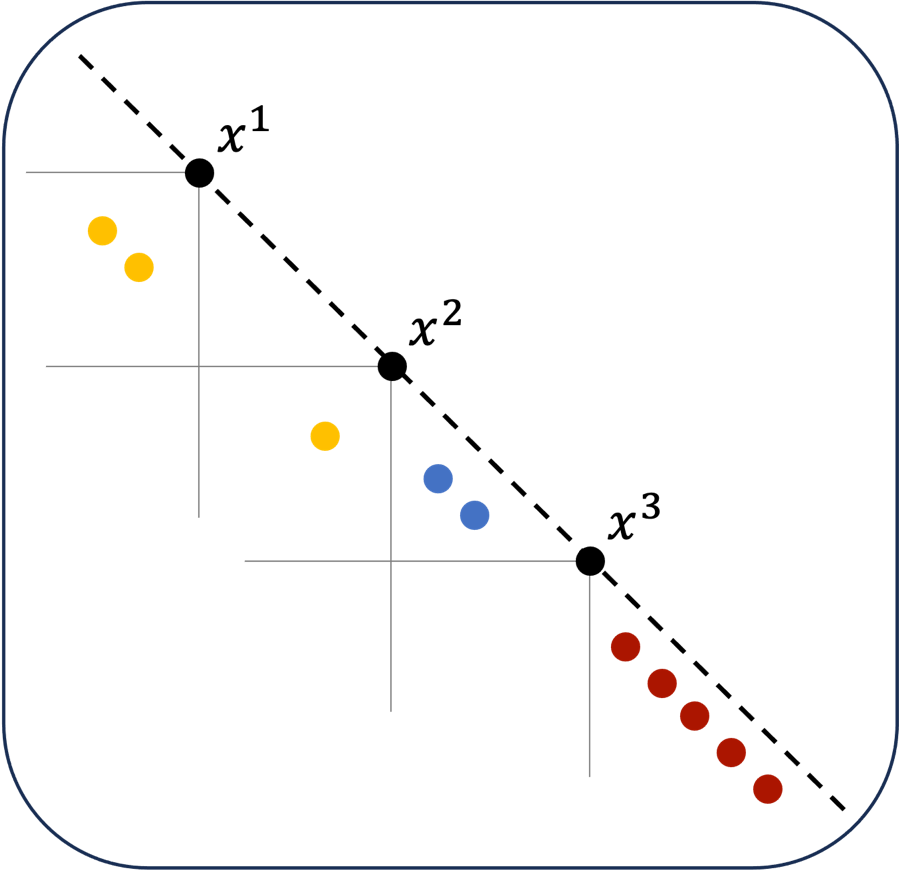}
	\caption{Set rank reversal}
	\label{fig:SetReversal}
	\end{minipage}
\end{figure}

The comparison of Figures \ref{fig:SetRanking} and \ref{fig:SetReversal} shows that rank reversal can occur with respect to the $R^\trido_C$-ranking even if only alternatives are added which are non-comparable to all existing ones. This is inline with the correspondding feature of the point ranking via $r_{X, C}$: adding the red points in Figure \ref{fig:SetReversal} to the situation in the right picture of Figure \ref{fig:SetRanking} produces $R^\trido_C(A^1) = \max\{5, 6\} = 6$, $R^\trido_C(A^2) = \max\{3, 6\} = 6$, $R^\trido_C(A^3) = \max\{3, 5\} = 5$ which reverses the ranking of $A^2$ and $A^3$ in these two situations.

The addition of alternatives which are comparable to some existing ones can also lead to rank reversal with respect to $\mathcal C_X$ whereas no rank reversal can occur when only alternatives are added which are not comparable to the existing ones.

This section is concluded with a discussion of $\mathcal C_X$, $R^\trido_C$ as unary indicators in the terminology of \cite{ZitzlerThieleBader10}. It follows directly from \cite[Thm. 3.1]{ZitzlerThieleBader10} and monotonicity that both $\mathcal C_X$ and $R^\trido_C$ generate weak refinements of the order relation $\succeq_C$ by
\begin{align*}
A \succeq_{\mathcal C_X} B \quad & :\Leftrightarrow \quad \mathcal C_X(A) \geq \mathcal C_X(B) \\
A \succeq_{R^\trido_C} B \quad & :\Leftrightarrow \quad R^\trido_C(A) \geq R^\trido_C(B).
\end{align*}
It remains to be clarified if these indicators are also refinements, i.e., if $A \succeq_C B$, $B \not\succeq_C A$ implies $\mathcal C_X(A) > \mathcal C_X(B)$ and parallel for $R^\trido_C$. For definitions, compare \cite[Def. 2.4, 2.5 and Thm. 3.1]{ZitzlerThieleBader10}. 

\begin{proposition}
\label{PropRefinements}
The relation $\succeq_{\mathcal C_X}$ is a refinement for $\succeq_C$ on $\mathcal P(X)$. The relation $\succeq_{R^\trido_C}$ is a weak refinement for $\succeq_C$ on $\mathcal P(X)$, but not a refinement in general.
\end{proposition}

{\sc Proof.} If $A, B \subseteq X$, then $A \succeq_C B$ implies $\{x \in X \mid \exists a \in A \colon a \geq_C x\} \supseteq \{x \in X \mid \exists b \in B \colon b \geq_C x\}$. On the other hand, $B \not\succeq_C A$ means that there is $\bar a \in A$ such that $b \not\geq_C \bar a$ for all $b \in B$. Thus, one has $\bar a \in \{x \in X \mid \exists a \in A \colon a \geq_C x\}$ since $A \subseteq X$ and $\bar a \not\in \{x \in X \mid \exists b \in B \colon b \geq_C x\}$ which means $\mathcal C_X(A) \geq \mathcal C_X(B) + 1$.

Already the situation depicted in the left graph of Figure \ref{fig:SetRanking} with $B = \{x^2, x^3\}$ and $A = \{x^1\} \cup B$ shows that $A \succeq_C B$, $B \not\succeq_C A$ do not imply $R^\trido_C(A) > R^\trido_C(B)$.

\smallskip
The monotonicity property in Proposition \ref{PropSetRankMonotonicity} is stronger than the weak refinement property in \cite[Def. 2.4, Thm. 3.1]{ZitzlerThieleBader10}. This shows that the relation $\succeq_{R^\trido_C}$ is somehow in between a weak refinement and a refinement. The examples in Figures \ref{fig:SetRanking} and \ref{fig:SetReversal} also make it clear that $\succeq_{R^\trido_C}$ reacts to the addition of non-comparable points while $\succeq_{\mathcal C_X}$ does not. Finally, let us mention that according to \cite[Thm. 3.1]{ZitzlerThieleBader10} a Richter-Peleg utility representation of $\succeq_C$, a popular concepts in economics (see, for example, \cite{AlcantudBosiZuanon16} and the references therein), is a refinement in the sense of \cite[Def. 2.4]{ZitzlerThieleBader10}; apparently, also this connection has never been made before in the MCDM literature.

\section{Classification models}

A basic question often is: how to find the $\alpha$\% best alternatives in a set of $N$ alternatives? It is possible to use the level sets 
\[
L_{X,C}(n) = \{z \in \R^d \mid r_{X,C}(z) \geq n\},
\]
$n \in \N$, for this purpose. These level sets correspond to the set-valued cone quantiles introduced in \cite{HamelKostner18}. One can ask which is the greatest number $n$ satisfying $\#\{x \in X \mid x \in L_{X,C}(n)\} \geq \frac{N\alpha}{100}$ which even gives some information about the set $X$ of alternatives as a whole: the higher this number is (in relation to the number of alternatives), the bigger is the group of ``good'' alternatives.

\smallskip
{\bf Unsupervised classification.} A related questions is how to cluster the alternatives in the ``really good ones," ``the really bad ones" and ``the ugly ones" where the latter category is meant to include those alternatives which considerably deviate from the rest in a good way w.r.t. some of the criteria and in a bad way w.r.t. other criteria. For this purpose, one can consider the sets $L_{X,C}(N/2)$ (the good ones, i.e., with a ranking of at least $N/2$), $L_{X,-C}(N/2)$ (the bad ones since the direction of the cone is reverted) and $X \backslash(L_{X,C}(N/2) \cup L_{X,-C}(N/2))$. The set $L_{X,C}(N/2)$ corresponds to the multivariate median from \cite{HamelKostner18}. It should be noted that the sets $L_{X,C}(N/2)$, $L_{X,-C}(N/2)$ could even be empty (examples can be found in \cite{HamelKostner22}).

Again, the shape of these sets also gives information on the set of alternatives as a whole and can be utilized, for example, for machine learning procedures such as unsupervised classification models as defined in \cite{HastieEtAl09}. Such models take the cone $C$ and the parameter $n$ as an input in order to partition the data. In this case, the cone $C$ can be seen as an externally given ordering cone reflecting the preference of a decision maker and $n$ defines how strict this preference is applied to the data. The output of this method are the three labels discussed above: good, bad, and ugly, the latter one a new concept reflecting the presence of non-comparable alternatives.

\smallskip
{\bf Supervised classification.} Reversing the question, a supervised classification model can be developed by finding the set $L_{X,C}(n)$ that achieves the lowest error rate if the data $X$ is labeled and the cone $C$, the parameter $n$, or both are not given as an input. 

The following recommender system can serve as blueprint where the cone $C$ is given and the parameter $n$ must be determined: the alternatives in $L_{X,C}(n)$ will be recommended where the cone $C$ stems from the given preference relation of the customer/user. The goal is to find, by adjusting $n$, the set $L_{X,C}(n)$ that includes the alternatives that will be accepted (with a high probability). In this case, the alternatives which are represented by their attributes, get a label that provides the information whether or not they are acceptable. The goal is to determine the set $L_{X,C}(n)$ and the value $n$, respectively, that minimizes the classification error rate. By comparing this implementation of $L_{X,C}(n)$ with the logistic regression, as described in \cite{JamesEtAl13}, two observations can be made. Firstly, the logistic regression has a response variable/label $Y$ and a predictor $X$ as input, whereas the new method has the preference information $C$ as additional input. The preference information can be seen as an additional tuning/customization of the model. Secondly, the parameter $n$ can also be seen as a threshold (probability cut-off value) as it affects the classification into the labels.

The further development into a semi-supervised model is straightforward as the unlabeled data can be assigned to the labeled one via the order relation $\leq_C$. If there are unlabeled alternatives dominating already positively labeled ones w.r.t. $\leq_C$, then these are also labeled as acceptable. The same principle is applied in the opposite direction for undesirable data. If the amount of unlabeled data is smaller then the labeled one and some of the former data is not comparable with the latter one, then this unlabeled data can be omitted for the calculation of the set $L_{X,C}(n)$ and reinserted as new data later on. The challenge is to create an algorithm that determines the set $L_{X,C}(n)$ and the value $n$, respectively, that minimizes the classification error rate. The false positives as well as false negatives should be minimized. This could be achieved by adjusting $n$ via appropriate methods. The resulting set $L_{X,C}(n)$ is then used to classify new data points. Of course, this set can be updated after a specified threshold of new data.

\begin{figure}[h]
	\centering
	\begin{minipage}{.6\textwidth}
	\includegraphics[width=.9\textwidth]{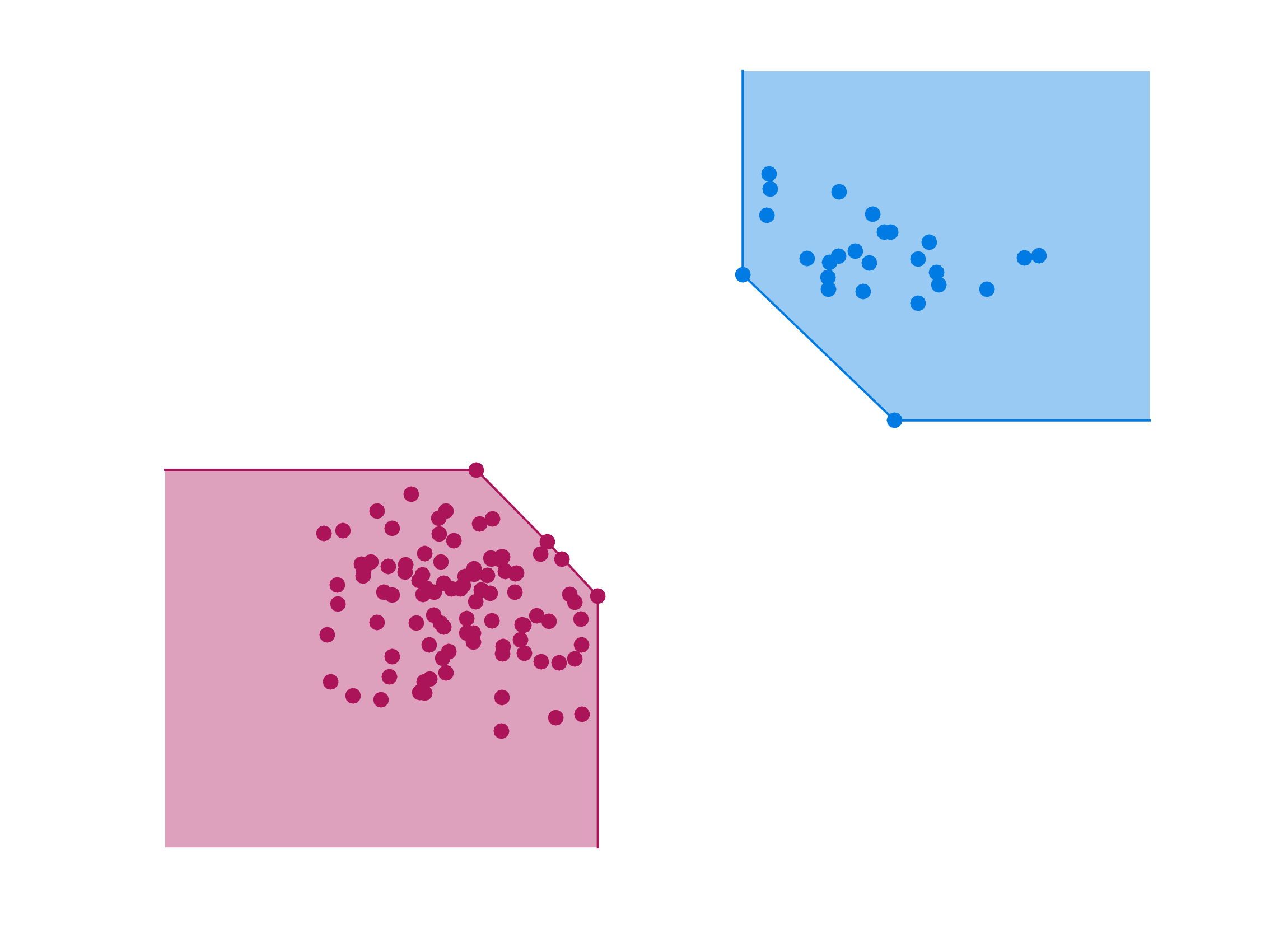}
	\caption{$X$ and $C$ are aligned.}
	\label{fig:aligned}
	\end{minipage}
\end{figure}

\begin{figure}[h]
	\centering
	\begin{minipage}[b]{.3\textwidth}
	\centering
	\includegraphics[width=1.1\textwidth]{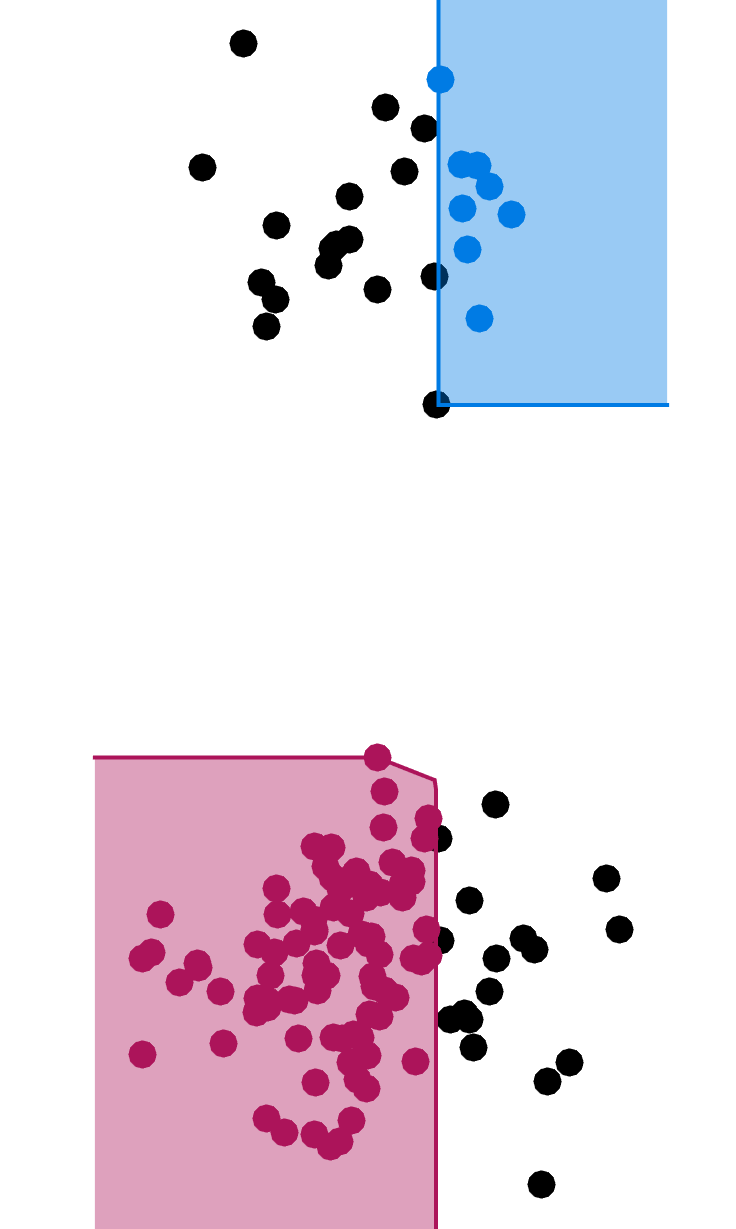}
	\caption{$X$ and $C$ are not aligned.}
	\label{fig:not_aligned}
	\end{minipage}
	\hspace{4cm}
	\begin{minipage}[b]{.3\textwidth}
	\centering
	\includegraphics[width=1.1\textwidth]{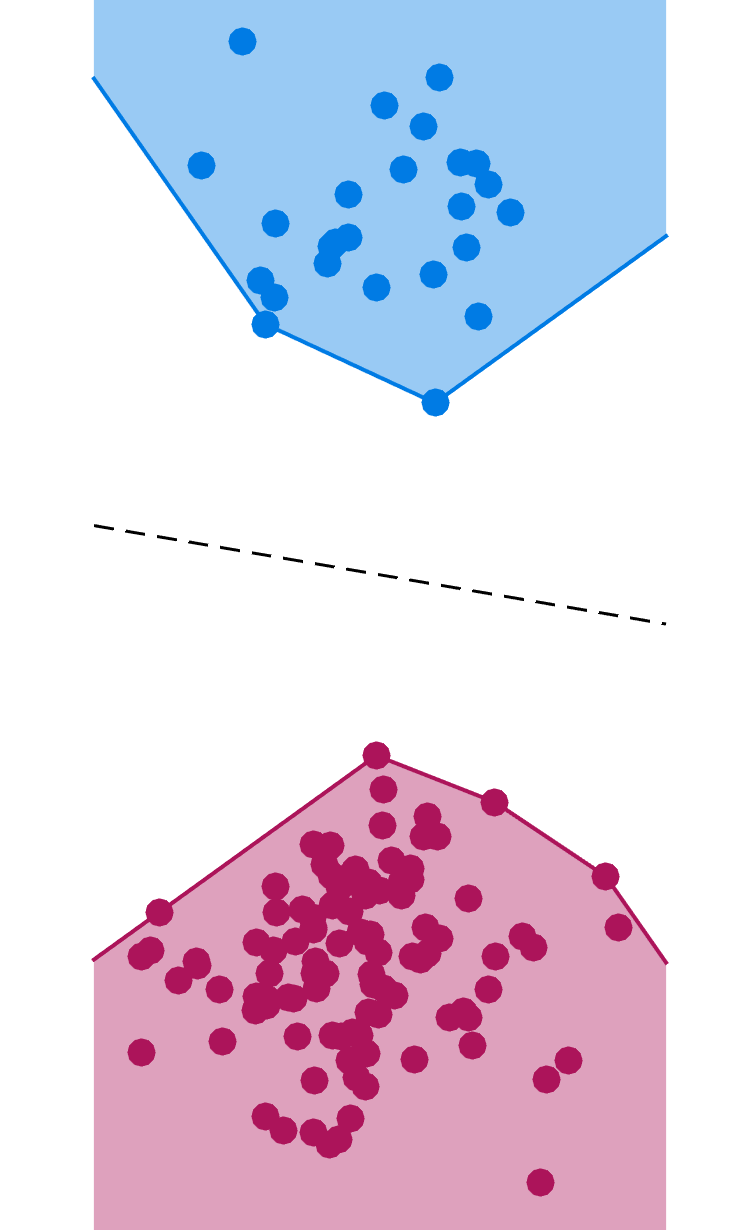}
	\caption{Alignment of $C$ via SVM}
	\label{fig:aligned_svm}
	\end{minipage}
\end{figure}

\smallskip
{\bf Towards preference learning.} If the cone $C$ is understood as a model for a customer preference, one might want to learn about it \cite{FurnkranzHullermeier10}. If there are two classes inside $X$, the parameter $n$ can be extrapolated from the number of elements in each class. This simple method works especially well if the cone $C$ is ``aligned'' with the data $X$ producing level sets with low classification error rates: compare Figure \ref{fig:aligned}. However, if the cone $C$, seen as a first estimate of the customer preference, is not aligned with the data $X$, the level sets could not include all relevant alternatives resulting in poor classification, see Figure \ref{fig:not_aligned}. This could be resolved by rotating the cone $C$ towards the data $X$.

The next idea not only looks for an optimal parameter $n$, but also partially adjusts the cone $C$ to the data. The linear Support Vector Machine (SVM) is used to align the cone $C$ with the data. The SVM separates the classes with a maximal margin hyperplane that. The idea is to adjust the cone $C$ to the data $X$ by rotating it towards the normal vector of the maximal margin hyperplane $w_{SVM}$ as in Figure \ref{fig:aligned_svm}. 

In higher dimensions ($d > 2$) this can be an elaborated procedure. For the computation of $L_{X,C}(n)$ the vertex representation of the dual cone $C^+$ is needed. This can be used to find an interior direction $w_{int}$ of $C^+$ by taking the convex combination of the extreme directions of $C^+$. Now, the rotation matrix from $w_{int}$ to $w_{SVM}$ is calculated. This matrix is used to rotate the dual cone $C^+$ towards $w_{SVM}$. This method only works if the two classes can be clearly separated by a hyperplane. This process can also be interpreted as a refinement of a classification via the SVM by introducing the order (preference) given by the ``data adjusted'' cone $C$.

Of course, the cone $C$ can be adapted to the data not only via a rotation, but also by changing its general shape. However, already in three dimensions, there are infinitely many degrees of freedom to change the cone. We consider this question a very interesting challenge since it gives a new direction of research linking the analysis of ``mere" data with preference relations. It brings the statistical concepts from \cite{HamelKostner18}, especially the level sets $L_{X,C}(n)$, to the field of preference learning as presented, e.g., in \ref{FurnkranzHullermeier10}.

\section{Conclusion and perspective}

A new ranking function is proposed which is derived from a statistical function \cite{HamelKostner18} which in turn resembles and generalizes the so-called half-space or Tukey depth function.  In statistics, a variety of more recently introduced depth functions with different properties exist. The ranking function proposed in Definition \ref{DefConeRanking} is based on a cone version of the half-space depth function. The same idea can be applied to other statistical depth functions like the zonoid depth \cite{Mosler02} or expectile depth functions \cite{HaHamel23}. This may lead to a variety of new ranking functions for MCDM. The computation of the values of $r_{X, C}$ is a non-trivial task but can be done based on a merge of sorting algorithms with convex geometry methods, compare \cite{HamelKostner22} for first impressions and remarks on complexity. The proposed ranking methods give a ranking of a given set of alternatives relative to each other. Therefore, rank reversal features appear naturally, but can be characterized and used in order to analyse the decision making procedure itself. Extensions to set preferences open the path to completely new comparison methods, e.g., for evolutionary multiobjective optimization on the one hand and new scalarization methods for multi-criteria optimization problems on the other hand. Data analysis for ordered data via statistical learning procedures are within reach based on the concepts in this paper.


\end{document}